\documentclass[letterpaper, 10 pt, conference]{ieeeconf}  

\IEEEoverridecommandlockouts                              

\usepackage{xcolor}
\usepackage{graphicx}
\usepackage{dsfont} 
\usepackage{amsmath,amsbsy,amssymb,mathtools}
\usepackage{gensymb}
\usepackage{bbold}          
\usepackage{nicefrac}       
\usepackage{microtype}      
\usepackage{tikz}
\usepackage{tabularray}

\usepackage{url}
\usepackage{color}
\usepackage{subcaption}
\usepackage[utf8]{inputenc}
\usepackage[T1]{fontenc}
\usepackage{balance}
\usepackage{cite}
\usepackage{siunitx}

\usepackage{colortbl}
\definecolor{orange}{rgb}{1,0.5,0}

\title{\LARGE \bf
RAMP: A Risk-Aware Mapping and Planning Pipeline for Fast Off-Road Ground Robot Navigation
}

\author{Lakshay Sharma$^{1}$, Michael Everett$^{1}$, Donggun Lee$^{1}$, Xiaoyi Cai$^{1}$,  Philip Osteen$^{2}$, and Jonathan P. How$^{1}$
\thanks{$^{1}$Aerospace Controls Laboratory, Massachusetts Institute of Technology, 77 Massachusetts Ave., Cambridge, MA 02139, USA
        {\tt\small \{lakshays, mfe, donggun, xyc, jhow\}@mit.edu}}%
\thanks{$^{2}$DEVCOM Army Research Laboratory, Aberdeen Proving Ground, MD 21005, USA
        {\tt\small philip.r.osteen.civ@army.mil}}%
}

\begin{document}

\maketitle
\thispagestyle{empty}
\pagestyle{empty}

\renewcommand{\baselinestretch}{.93}
\begin{abstract}
A key challenge in fast ground robot navigation in 3D terrain is balancing robot speed and safety. Recent work has shown that 2.5D maps (2D representations with additional 3D information) are ideal for real-time safe and fast planning. However, the prevalent approach of generating 2D occupancy grids through raytracing makes the generated map unsafe to plan in, due to inaccurate representation of unknown space. Additionally, existing planners such as MPPI do not consider speeds in known free and unknown space separately, leading to slower overall plans. The RAMP pipeline proposed here solves these issues using new mapping and planning methods. This work first presents ground point inflation with persistent spatial memory as a way to generate accurate occupancy grid maps from classified pointclouds. Then we present an MPPI-based planner with embedded variability in horizon, to maximize speed in known free space while retaining cautionary penetration into unknown space. Finally, we integrate this mapping and planning pipeline with risk constraints arising from 3D terrain, and verify that it enables fast and safe navigation using simulations and hardware demonstrations.

\end{abstract}

\section{Introduction}

\begin{figure}
\centering
\begin{tabular}{cc}
    \includegraphics[trim = 0mm 0mm 0mm 0mm, clip, height=0.175\textheight]{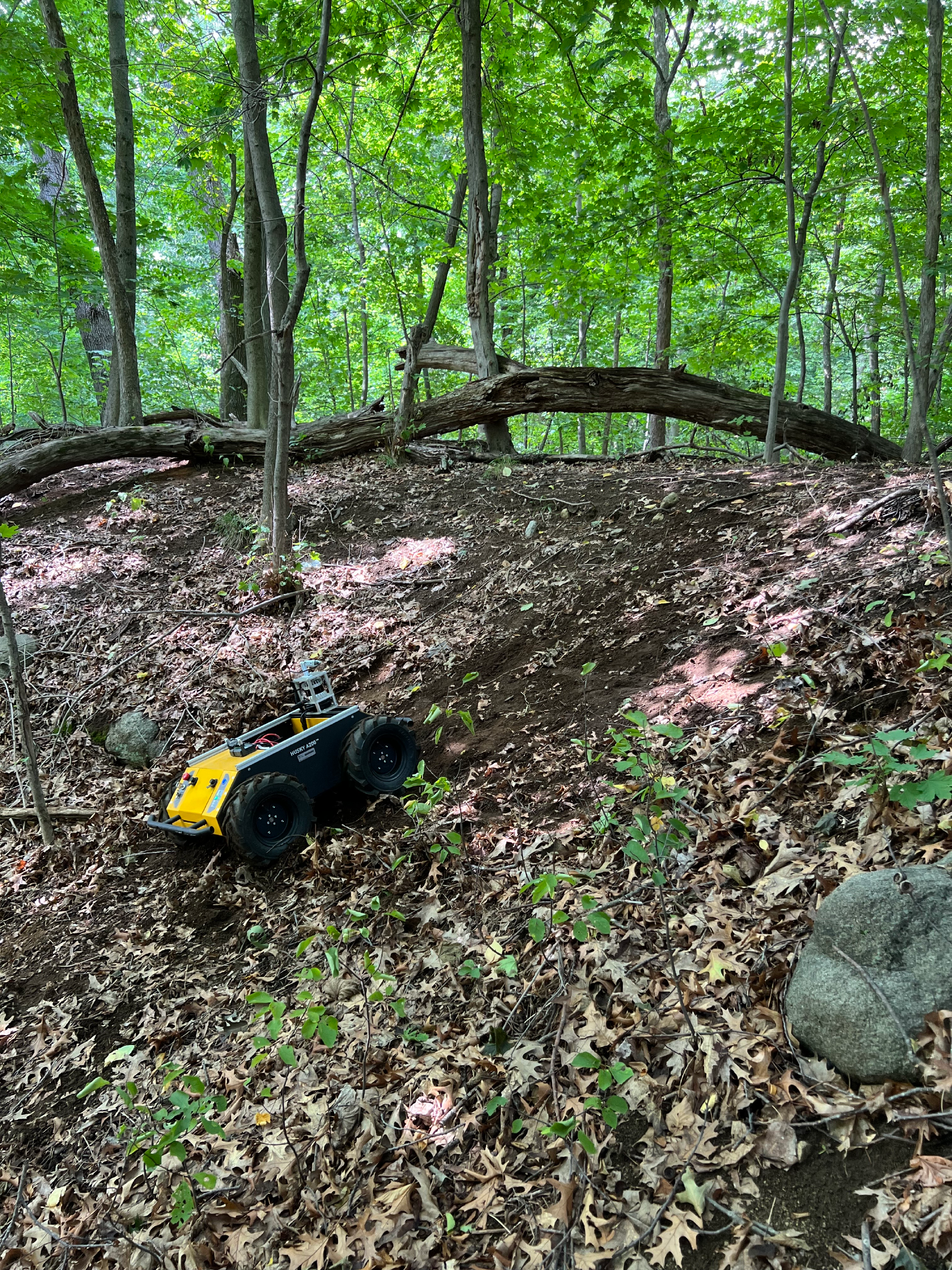} &
    \includegraphics[trim = 0mm 0mm 0mm 0mm, clip, height=0.175\textheight]{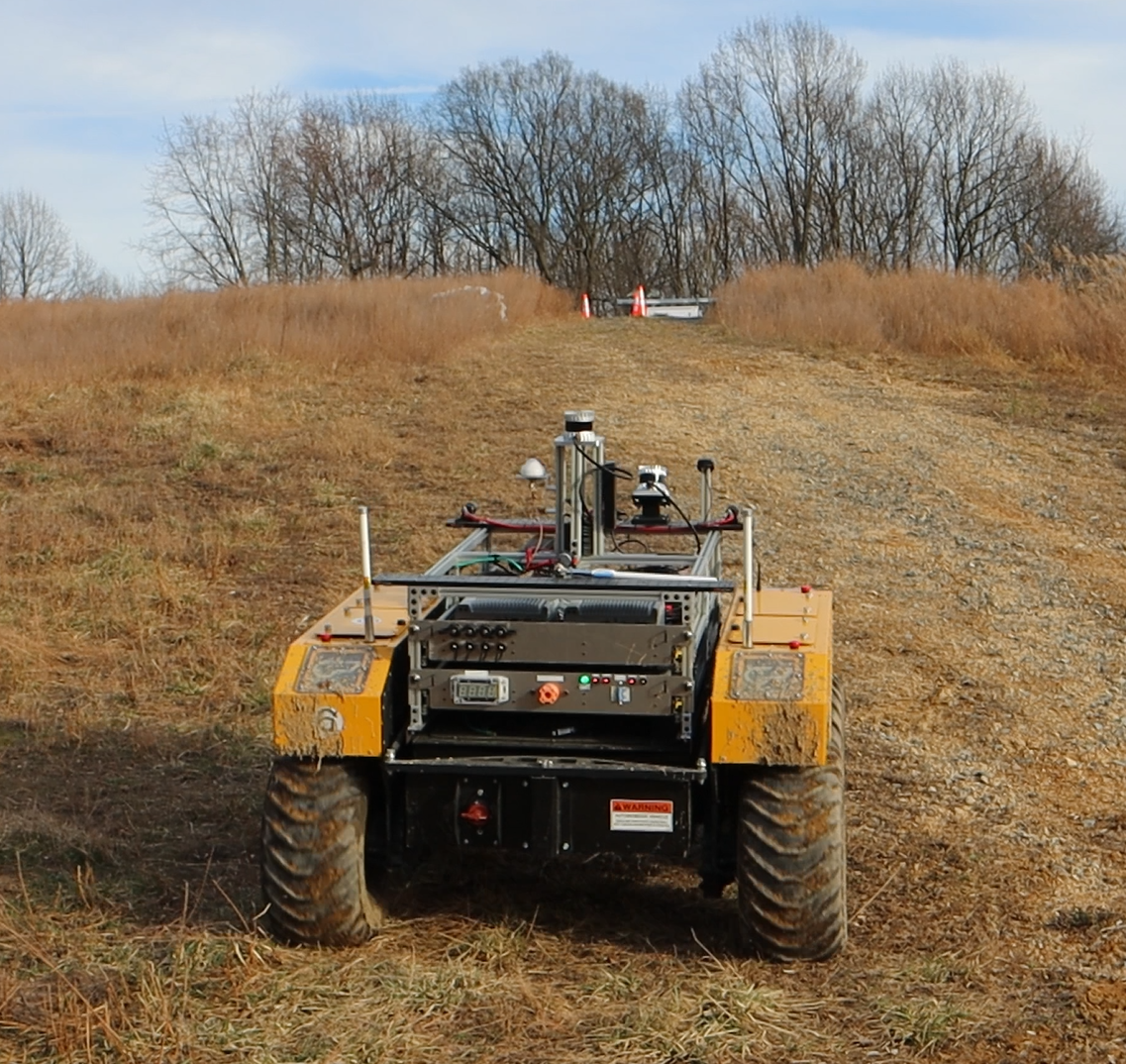} \\
    (a) & (b)
\end{tabular}
\caption{Ground robots (Clearpath Husky and Warthog) navigating 3D terrain.\label{fig:envs}}
\end{figure}

\begin{figure}[t!]
    \centering
    \includegraphics[trim = 0mm 0mm 0mm 0mm, clip, width=0.45\textwidth]{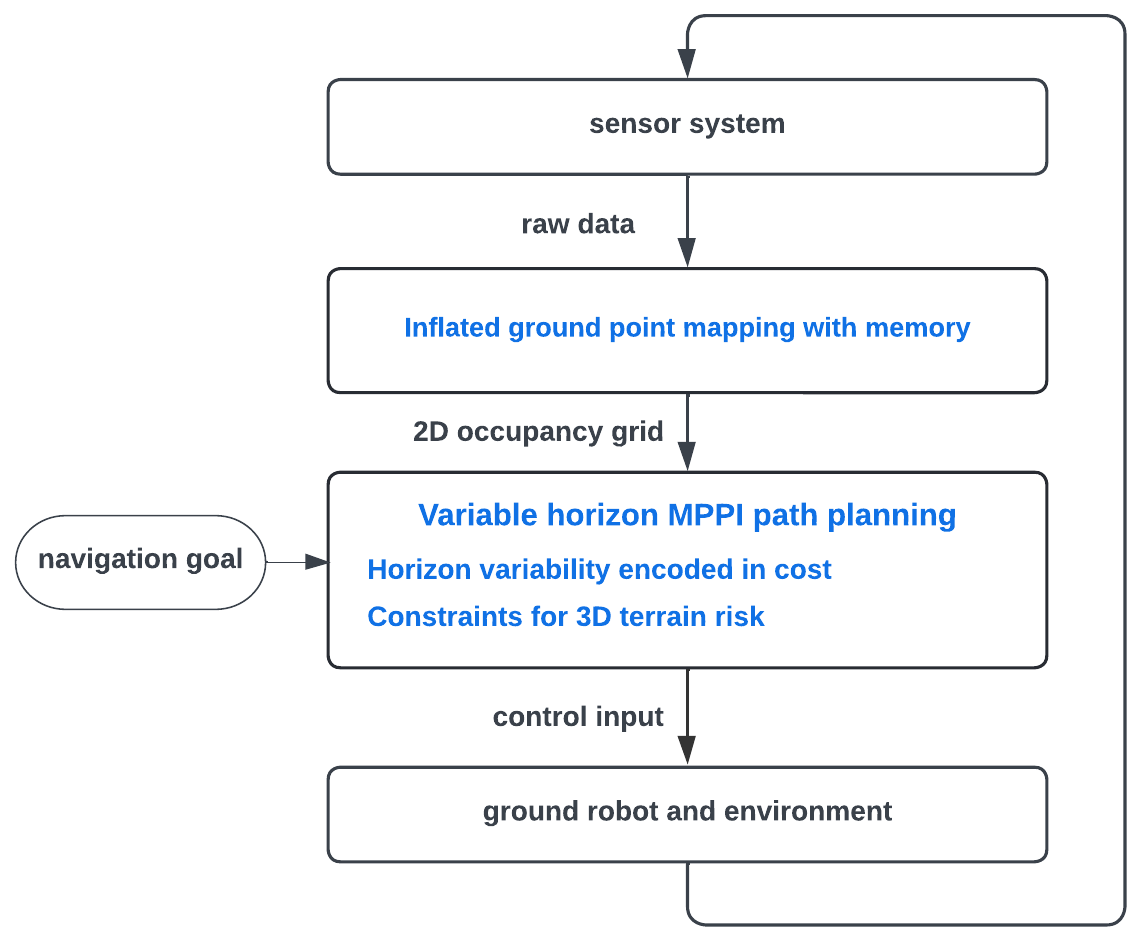}
    \caption{ This diagram illustrates our risk-aware mapping and planning pipeline (RAMP) for fast off-road ground robot navigation. (As implemented, the entire pipeline is real-time at 20 Hz) \label{fig:pipeline}}
\end{figure}

Ground robot navigation in outdoor 3D terrain is a longstanding area of research, with challenges compounded by the need for high speed as mobile robot platforms become increasingly capable. Some emergent time-critical applications include planetary exploration \cite{bares1989ambler,massari2004autonomous,voosen2018nasa}, search-and-rescue missions \cite{kantor2003distributed}, and combat missions \cite{yamauchi2004packbot}. This paper aims to develop fast off-road navigation for ground robots on unmapped 3D terrain, as shown in Fig. \ref{fig:envs}, by presenting a 2.5D local planning pipeline, as shown in Fig. \ref{fig:pipeline}.

\noindent\textbf{Variable horizon path planning for fast navigation:} 
A common representation of the environment for robot navigation involves classifying nearby space as known-free, known-occupied, or unknown regions, either in 2D (e.g., occupancy grid mapping~\cite{elfes1989using}), 2.5D (e.g., costmap2d (Robot Operating System, ROS plugin), multi-layer grid map~\cite{fankhauser2016gridmap}) or 3D (e.g., octomap~\cite{hornung2013octomap}, surfel representation~\cite{stuckler2014multi}, pointcloud map~\cite{wiemann2019file}, signed distance field~\cite{oleynikova2016signed}). For vehicles constrained to the ground, a 2.5D representation provides richer information than a purely 2D representation while maintaining computational tractability
(e.g., occupancy grid map \cite{elfes1989using}, octomap \cite{hornung2013octomap}).
As discussed in \cite{tordesillas2021faster}, a key challenge is how the planner should interpret this representation. Safety-guaranteed path planning methods \cite{tordesillas2021faster,fridovich2019safely,kousik2020safe,janson2018safe} heavily penalize entering unknown space, which can lead to conservative plans and unnecessarily low robot speed. Alternatively, treating unknown space as free allows for faster navigation, but could cause dangerous maneuvers resulting from high-speed unknown space penetration (such as being unable to stop before unexpected obstacles or going airborne over a negative slope). It is therefore essential to bridge the gap between high robot speed and safe plans into unknown space.


Some works attempt to address this by penalizing high speed in unknown space \cite{cai2022risk}. Yet, these works do not incentivize moving quickly in free space, leading to slow average speeds over the mission duration and undesired plan curvature due to fixed planning horizons. To alleviate this issue, we present a planner cost function with horizon variability to generate fast, unknown space-aware trajectories in real-time. Our cost function allows for variable horizons \cite{shekhar2012robust} not just for the terminal time to reach the goal but also for the time of each known to unknown space transition. For real-time computation, our implementation builds on model predictive path integral (MPPI) \cite{williams2017model,cai2022risk}, where sampling and parallel optimization are performed.

\begin{figure}[t!]
    \centering
    \includegraphics[trim = 0mm 0mm 0mm 0mm, clip, width=0.45\textwidth]{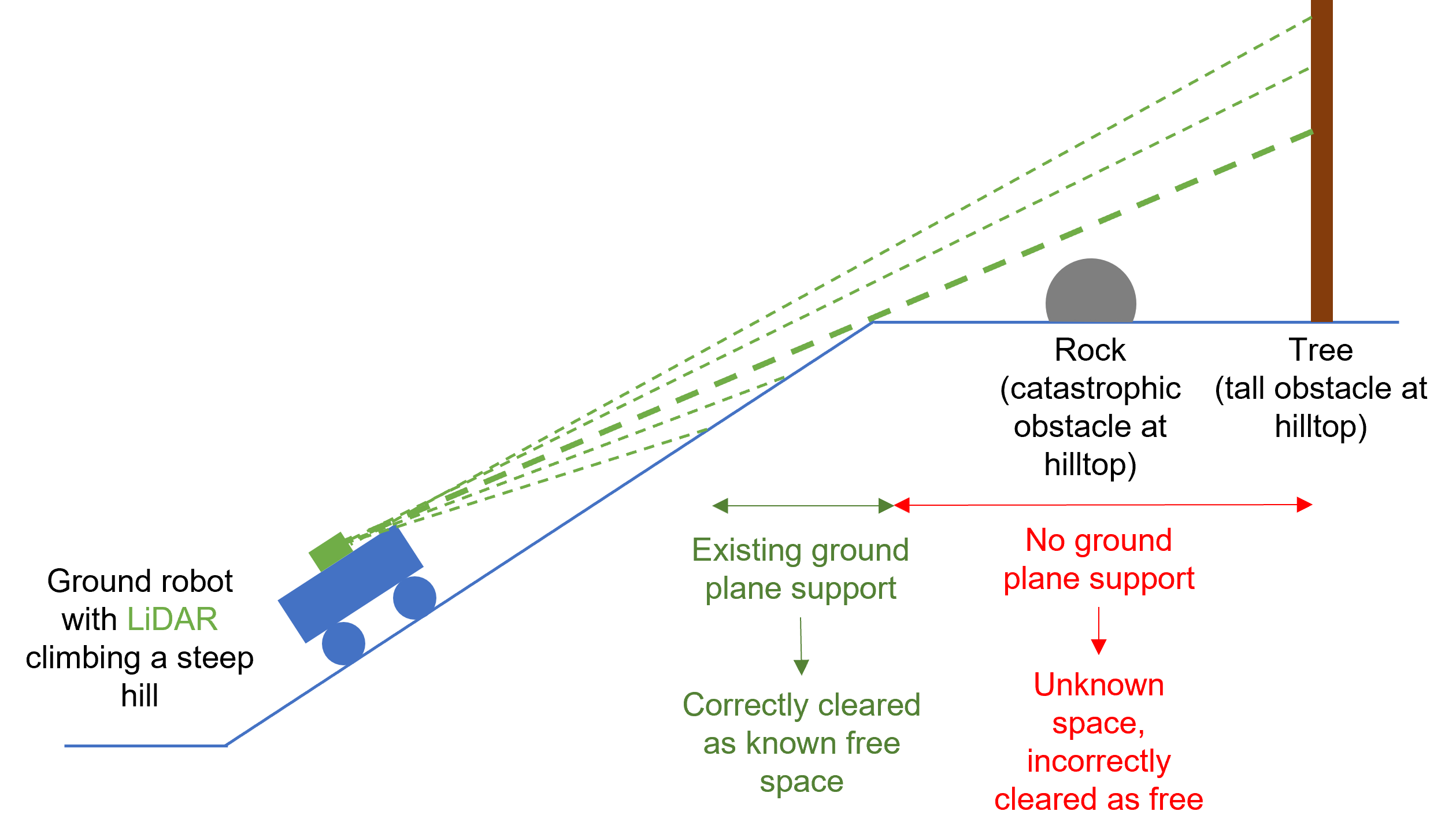}
    \caption{LiDAR raytracing leads to maps with inaccurate representations of unknown space. \label{fig:raytacing}}
\end{figure}

\noindent
\textbf{Ground point inflation for occupancy grid generation:} Since a ground robot is constrained to move on the ground plane, we make the choice to plan on a 2D map using 2D robot dynamics to save time on map access (direct for a 2D array vs. tree traversal for octomap \cite{hornung2013octomap}) and MPPI rollout state inference. A major challenge in using 2D maps for 3D environments \cite{zhou2020cylinder3d} is that the conventional method of raytracing shown in Fig. \ref{fig:raytacing} used to project LiDAR information onto an occupancy grid does not check for ground plane support. This leads to a map where not all known free space is necessarily traversable, thereby feeding false information to the planner. To have an accurate and efficient known vs. unknown space classification of the environment, we present ground point inflation as a direct way of generating a ground-plane-aware occupancy grid from semantically classified point clouds.

\begin{figure*}[th]
\centering
\begin{tabular}{cccc}
    \includegraphics[trim = 350mm 300mm 350mm 180mm, clip, height=0.132\textheight]{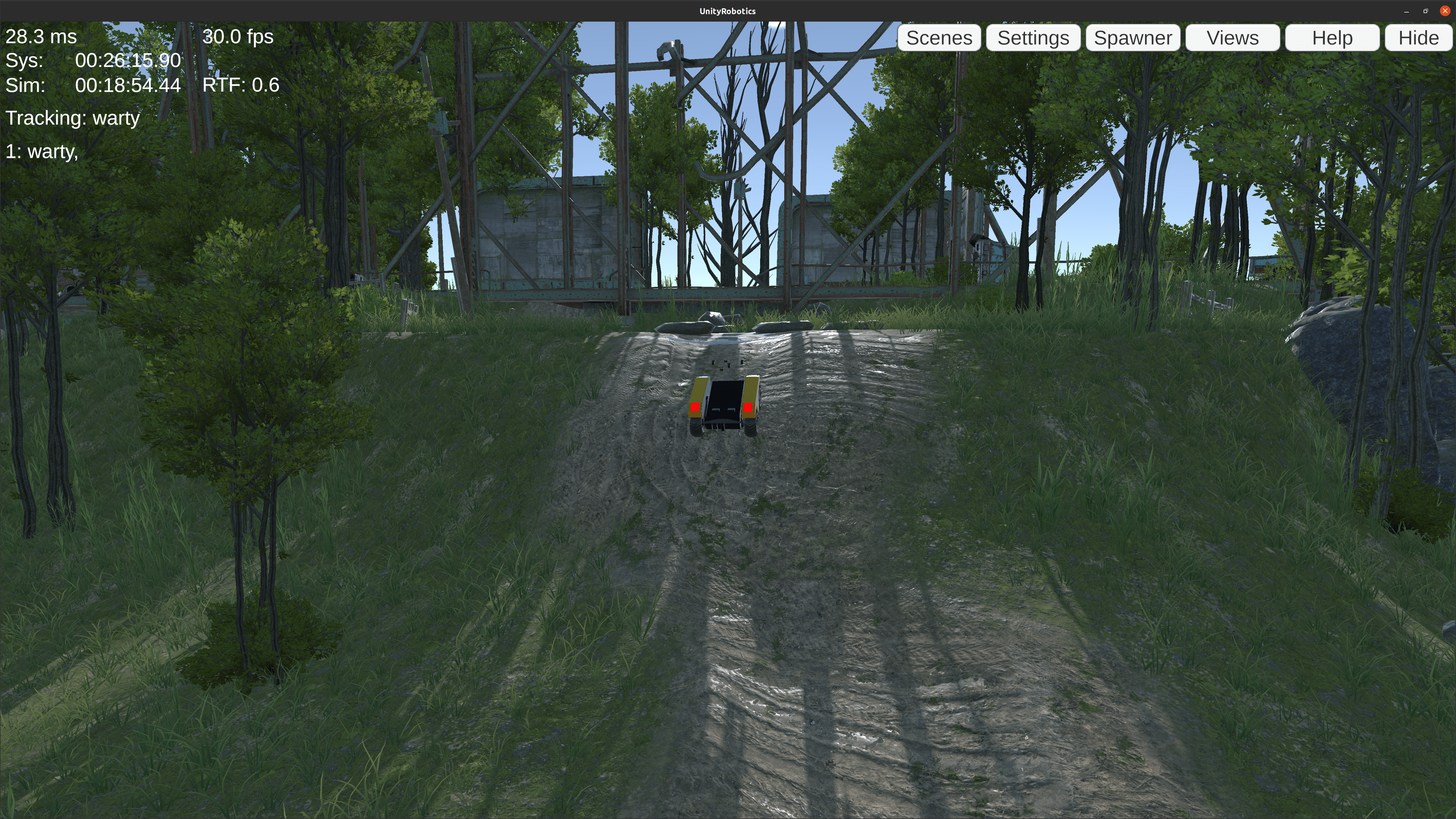} &
    \includegraphics[trim = 0mm 0mm 0mm 0mm, clip, height=0.132\textheight]{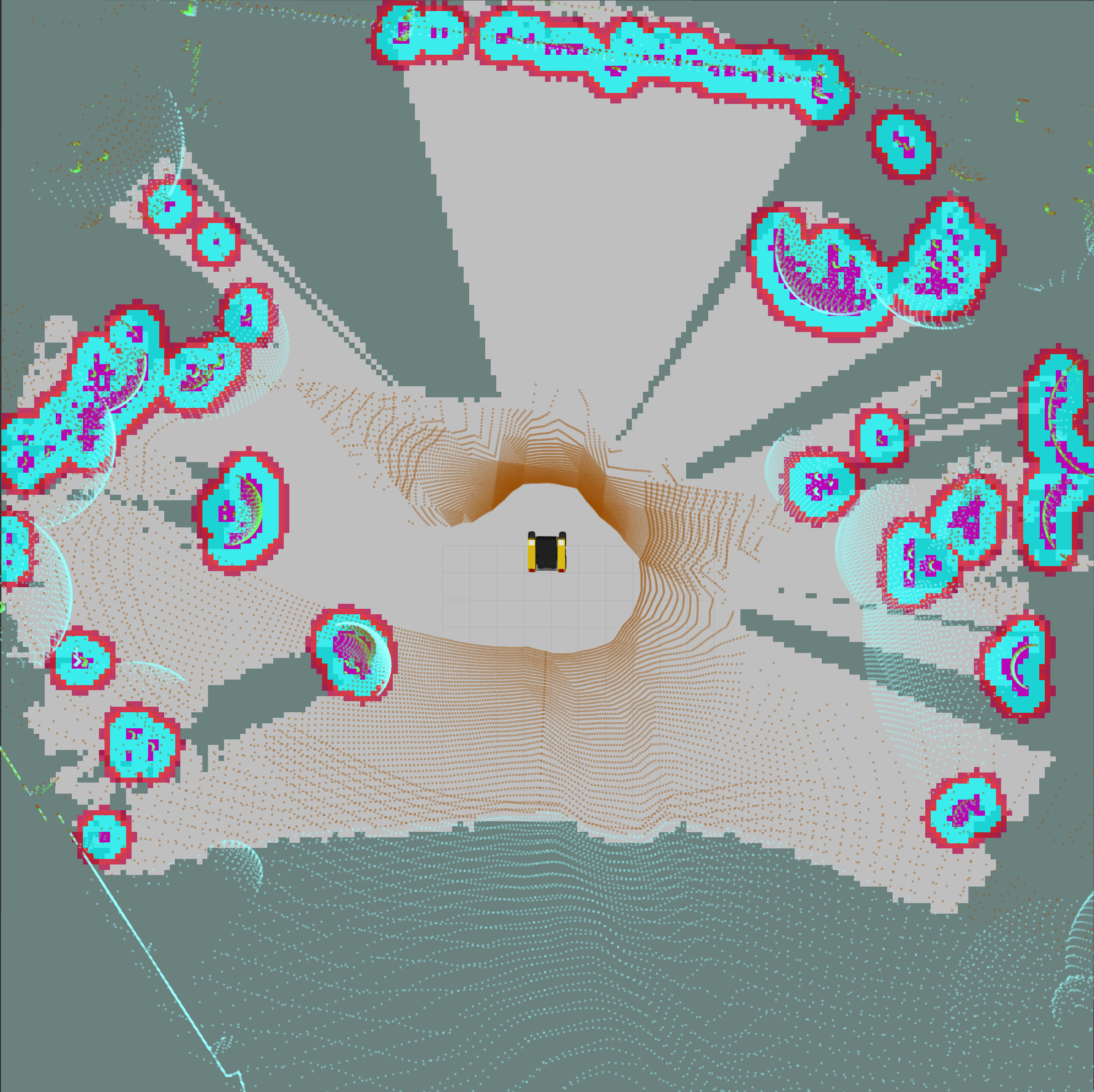} &
    \includegraphics[trim = 0mm 0mm 0mm 0mm, clip, height=0.132\textheight]{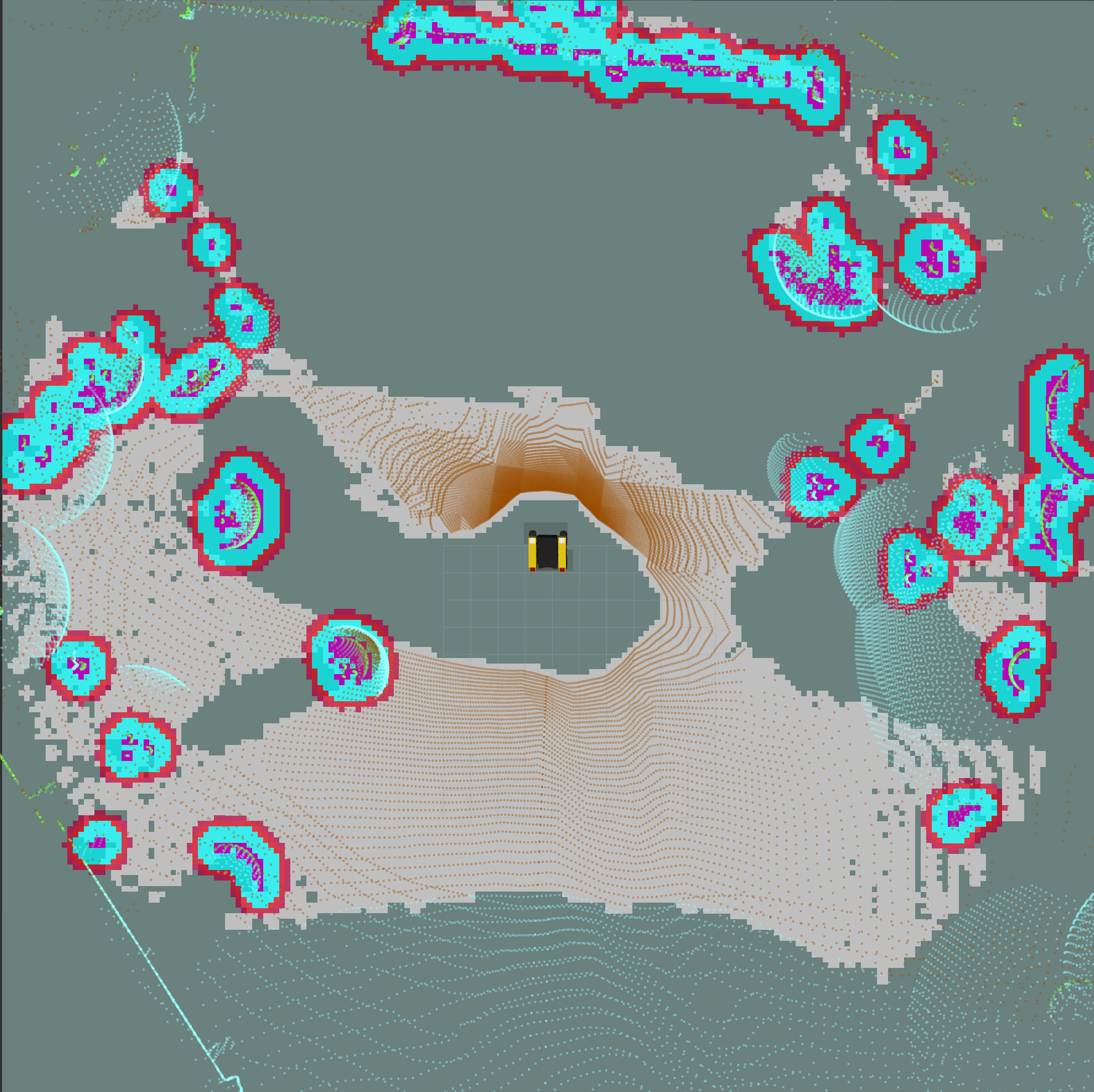} &
    \includegraphics[trim = 0mm 0mm 0mm 0mm, clip, height=0.132\textheight]{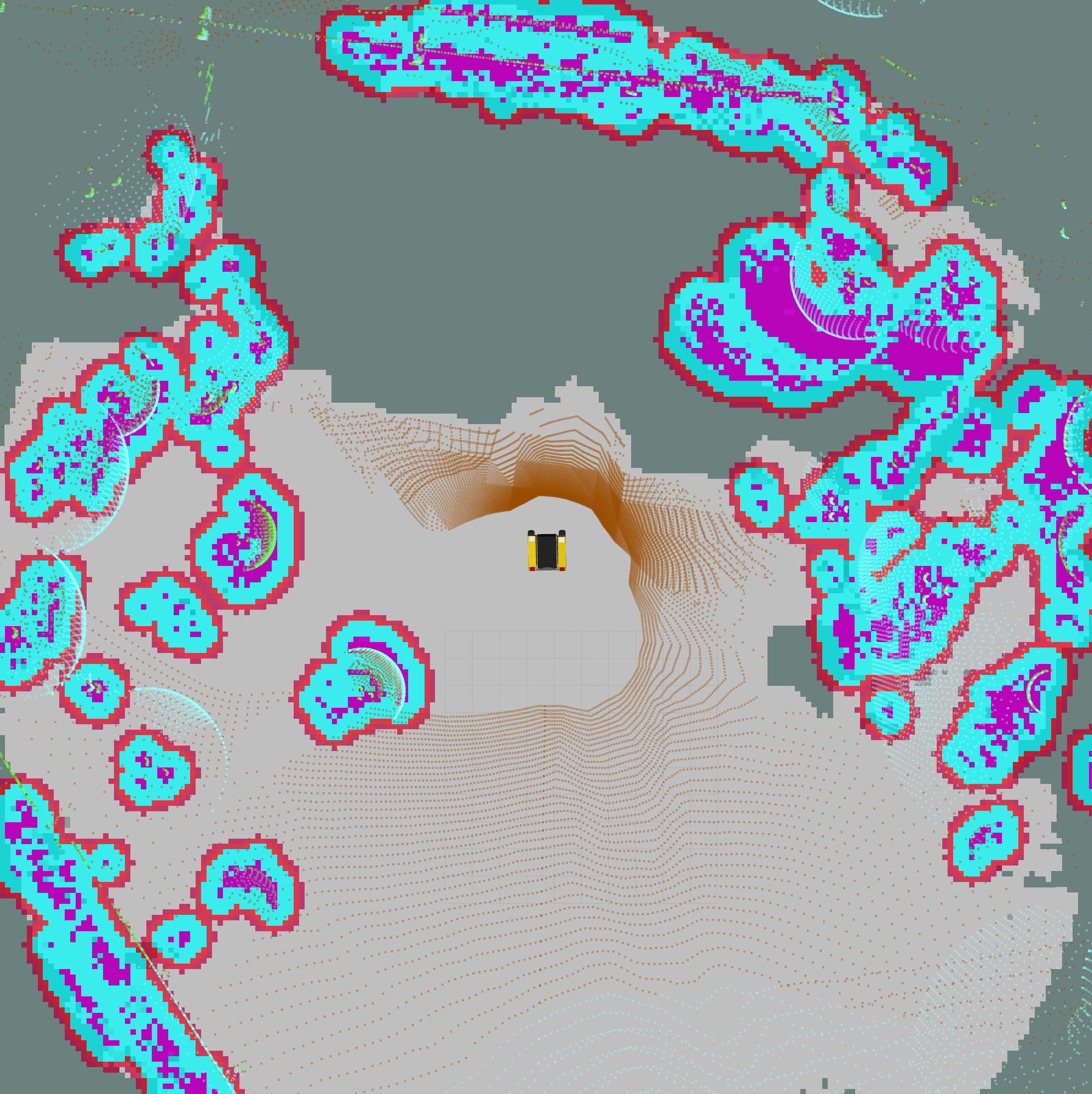} \\
    (a) & (b) & (c) & (d)
\end{tabular}
\caption{(a) Scenario in simulation, with the robot about to clear a hilltop with large rocks.  (b)-(d) Occupancy grids generated using: (b) Raytracing, (c) Ground point inflation without memory, (d) Ground point inflation with memory. The brown dots represent LiDAR hits classified as ground, light gray represents known free space, darker-gray represents unknown space, and the objects with a magenta to cyan gradient represent known occupied space. The rocks on the hilltop and the surface of the hilltop itself aren't detectable by LiDAR, but raytracing erroneously clears it as known free space. Meanwhile, memory is essential to have a usable map for the planner, such that the robot is not surrounded by unknown space.\label{fig:Map_with_without_memory}}
\end{figure*}

\noindent
\textbf{Planning constraints imposed by 3D terrain:} A fully 2D mapping/planning pipeline like the one described so far is incapable of dealing with risks arising directly from the 3D nature of the problem. These risks include the robot tipping over due to turning on a slope, climbing a steep slope, or taking tight turns at high speed. To round out our variable horizon planner, we add constraints for all three of these specific maneuvers, using information such as surface normal components and slope of the 3D terrain by employing the GridMap \cite{fankhauser2016gridmap} library. This makes the pipeline akin to 2.5D by incorporating 3D constraints in a 2D planner.

In summary, our contributions include:
\begin{itemize}
    \item A novel 3D-aware 2D map representation based on lidar point inflation that can accurately assess the 3D occupancy information that's relevant for driving (thus the risk induced by wrongly classified known space), while retaining the computational advantage of a 2D map structure;
    \item A risk-aware planner that uses variable time-horizon cost formulation to achieve high-speed navigation within known space and low speed in unknown space, where the speed in the unknown is adjustable based on risk-tolerance;
    \item A demonstration of our approach with proposed risk-aware planner and map representation simultaneously achieving higher success rate and time-to-goal than approaches with existing 2D mapping techniques and planners in a high-fidelty Unity simulator.
\end{itemize}

\section{Ground Point Inflation and Map Memory}
\label{sec:GroundPointInflationMapMemory}

Generating an occupancy grid for the planner from LiDAR data is the first step involved in our pipeline. Given a pointcloud classifier capable of labeling points as navigable (referred to as ground) or not (referred to as obstacle), all points classified as ground form the basis of ground plane support for the robot. Known free space in the 2D occupancy grid is then defined as the collection of circles of a fixed inflation radius around each ground point. This allows for a more accurate representation of unknown space, as regions without ground support information are not cleared as free.

The point cloud classifier used here is a simple geometric one that uses sharp increases in feature height as criteria for defining obstacles. However, several other methods exist that estimate semantic properties within the environment, such as Cylinder3D \cite{zhou2020cylinder3d} and SalsaNet \cite{aksoy2020salsanet}.
However, simply using this inflated ground point representation in its raw form also presents the following issues:
\paragraph{Map sparsity at long range} Due to finite sensor resolution, the distance between LiDAR beams increases with increasing distance from the robot. When this distance between neighboring beams (and therefore hits) exceeds the ground point inflation radius, known free space is no longer continuous, and the map generated from a single LiDAR scan, therefore, becomes sparse at long range. 

\paragraph{Tilt-induced map amnesia} Due to the uneven terrain of interest, any amount of robot tilt induces a nonuniform spread of LiDAR points. This leads to large parts of the map being classified as unknown space, which had previously been classified as known before the tilt. For large tilt values, entire sectors of the map around the robot might be classified as unknown space, even though they were known to be free before the tilt. For risk-aware planning into unknown space, it is not ideal to have such an inaccurate classification of known space as unknown.

\paragraph{LiDAR shadow} The 3D LiDAR sensors commonly used on robots today have a limited vertical field of view, so depending on sensor height, the closest LiDAR hits might not be in the immediate neighborhood of the robot. Additionally, some of the closest beams might have to be masked out in pointcloud preprocessing to remove LiDAR self-hits. This creates a region in the immediate vicinity of the robot that has no LiDAR hits and is defined here as LiDAR shadow. Since there are no points to be classified, it translates to a ring of unknown space around the robot in the occupancy grid.

We address all these issues by implementing persistent map memory: filling in patches of unknown space as the robot moves based on information about each patch in prior maps. This is done by first aggregating the raw maps from the last $t$ seconds, where $t$ can vary. Next, any unknown space in  the map is filled in with information about that location in the processed final map from the previous iteration of this memory pipeline: if it was previously known to be free or an obstacle, it is classified as such, otherwise, it is left as unknown space. This generates a much more densely populated and usable map for planning, as shown in Fig. \ref{fig:Map_with_without_memory}. 

\section{Variable Horizon Cost Function}
\label{sec:VariableHorizonCostFunction}

This section presents a variable-horizon optimal control formulation, which builds on \cite{cai2022risk}.
We consider a state trajectory solving
\begin{align}
    \textbf{x}_{t+1} = f(\textbf{x}_t, \textbf{u}_t),
    \label{eq:dynamics}
\end{align}
where $\textbf{x}_t\in\mathbb{R}^n$ is a state at time $t$, and $\textbf{u}_t\in\mathbb{R}^m$ is a control input at $t$. The state represents the robot's position and orientation (yaw), and the control represents the left and right wheel velocities, where we are considering a skid-steer dynamics model. 
We use $\textbf{x}_{0:T}$ to denote a state trajectory from $0$ to $T$.

\begin{figure}[t!]
    \centering
        \includegraphics[trim = 0mm 0mm 0mm 0mm, clip, width=0.6\columnwidth]{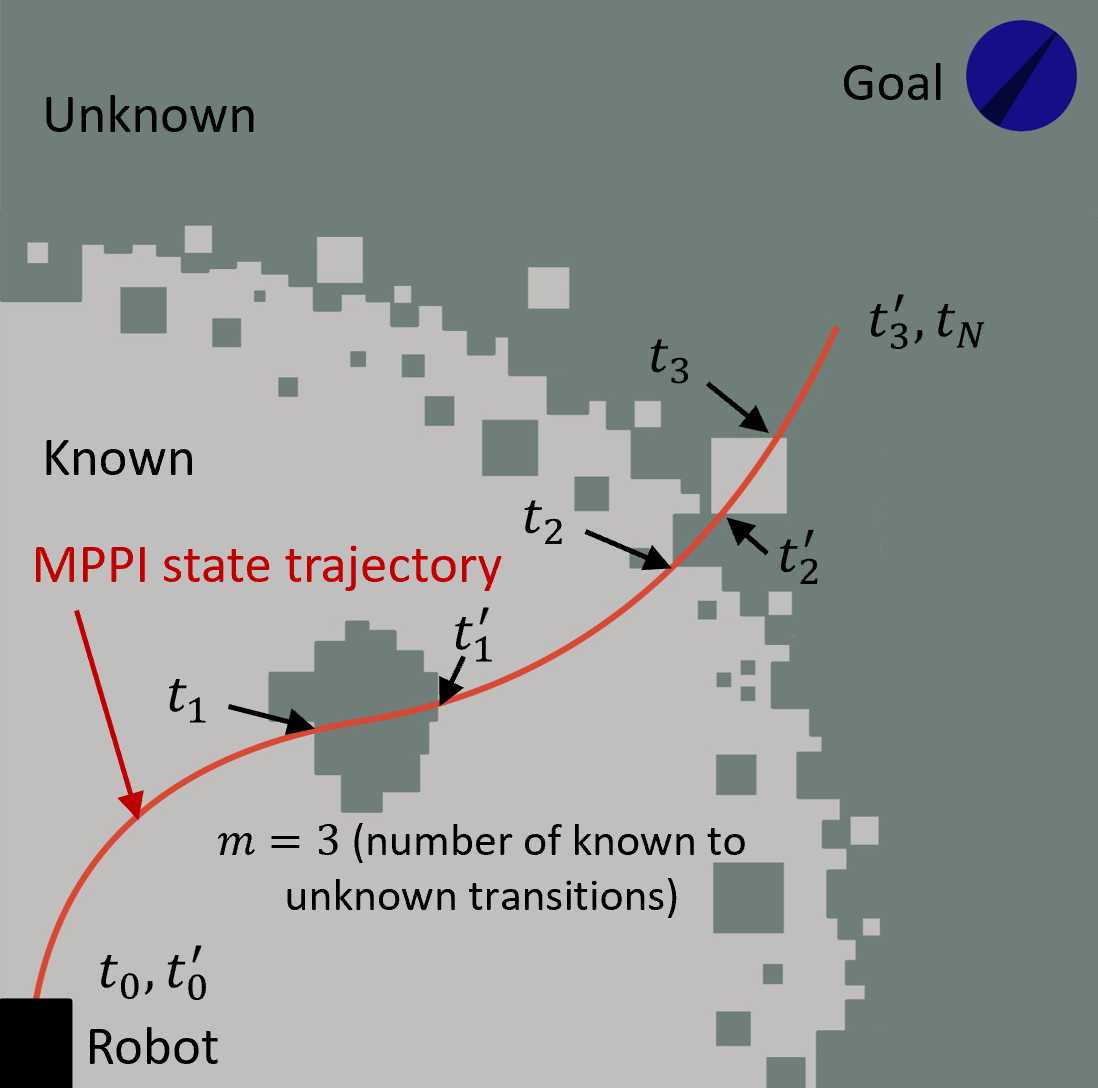}
    \caption{An illustration of the various horizons $t_i$ and $t_i'$ used in the cost function formulation. \label{fig:variable_horizon_notation}}
\end{figure}

We design a variable-horizon problem that enables the vehicle to move fast in known free space, slow down cautiously in unknown space, and prioritize goal approach as follows:
{\small
\begin{align}
    \min_{\textbf{u}_{0:T-1}} \frac{1-\beta}{m} \sum_{i=1}^m & \bigg[ \gamma(t_i - t'_{i-1})^2 + \phi(\textbf{x}_{t_i}) \bigg]\mathbb{1}\big(t_i \leq \text{TTR}(\textbf{x}_{0:T})\big) \notag \\
    &+ \beta\Big(\gamma \text{TTR}(\textbf{x}_{0:T})^2 + \phi(\textbf{x}_{\text{TTR}(\textbf{x}_{0:T})})\Big) \\
    &+  \alpha \frac{\sum_{i=1}^{m} \sum_{t=t_i}^{t'_{i}} \lVert \textbf{v}_t \rVert \mathbb{1}(t \leq \text{TTR}\big(\textbf{x}_{0:T})\big)}{\sum_{i=1}^{m} f(t_i' - t_i)}, \notag
\end{align} \label{eq:cost}}
where $\alpha(>0$ , usually around 60), $\beta \in [0.5,1]$ (usually set to 0.99), and $\gamma (>1$, usually around 50) are tunable coefficients, $m$ is the number of the state trajectory's transitions from known to unknown space within the planning horizon $T$ (10 seconds), $t_i$ is $i$-th transition time of the trajectory from known to unknown space, $t_i'$ is $i$-th transition time of the trajectory from unknown to known space, $f$ is the planner frequency (20 Hz),
$\textnormal{TTR}$ represents the time-to-reach within the planning horizon $T$
\begin{align}
    \textnormal{TTR}(\textbf{x}_{0:T}) = 
             \min\Big\{T, \min_{t\in\{0,...,T\}} \big[t~|~\textbf{x}_t \in \text{Goal}\big]\Big\}, 
\end{align}
where `$\text{Goal}$' is the navigation goal, 
$\mathbb{1}(\text{Condition})$ is an indicator function for the `Condition'
\begin{align}
    \mathbb{1}(\text{Condition}) = 
    \begin{cases}
        1, & \text{if the Condition holds},\\
        0, & \text{otherwise,}
    \end{cases}
\end{align}
and $\phi$ is the cost-to-go
\begin{align}
    \phi(\textbf{x}_t) = \left(\frac{\|\textbf{p}^{\text{goal}} - \textbf{p}_t\|}{s^{\text{default}}}\right)^2,
\end{align}
Here, $\textbf{p}^{\text{goal}}$ and $\textbf{p}_t$ are the positions of the goal and the robot at time $t$ respectively, and $s^{\text{default}}$ is the post-rollout speed used as a heuristic for the time-to-go. ($s^{\text{default}}$ is usually set to a low number such as 0.1 m/s to encourage goal approach)
Fig. \ref{fig:variable_horizon_notation} visually explains the notation $t_i$, $t'_i$, and $m$.

In the first term, $t_i-t'_{i-1}$ incentivizes the vehicle to go faster in free space (a source of horizon variability), and $\phi(\textbf{x}_{t_i})$ prevents the vehicle from greedily jumping to the nearest patch, instead pushing it to remain close to the goal at transition time. 
The second term results in the terminal state of the plan tending to approach the goal while minimizing the amount of time it takes to do so using $\text{TTR}$ explicitly (another source of horizon variability). 
The third term prevents high average speed in unknown space, ensuring plan safety for medium to high $\alpha$.

Overall, this cost function is able to provide us with the edge needed over default MPPI implementations due to the variability in goal and unknown space transition horizons, combined with unknown space speed cost.

\section{Risk Constraints}
\label{sec:RiskConstraints}

Fig. \ref{fig:risk_constraint} shows three kinds of risks that need to be accounted for, by constraining the cost function:
\begin{figure}[t!]
\centering
\begin{subfigure}{0.45\columnwidth}
    \includegraphics[trim = 0mm 0mm 0mm 0mm, clip, width=1\columnwidth]{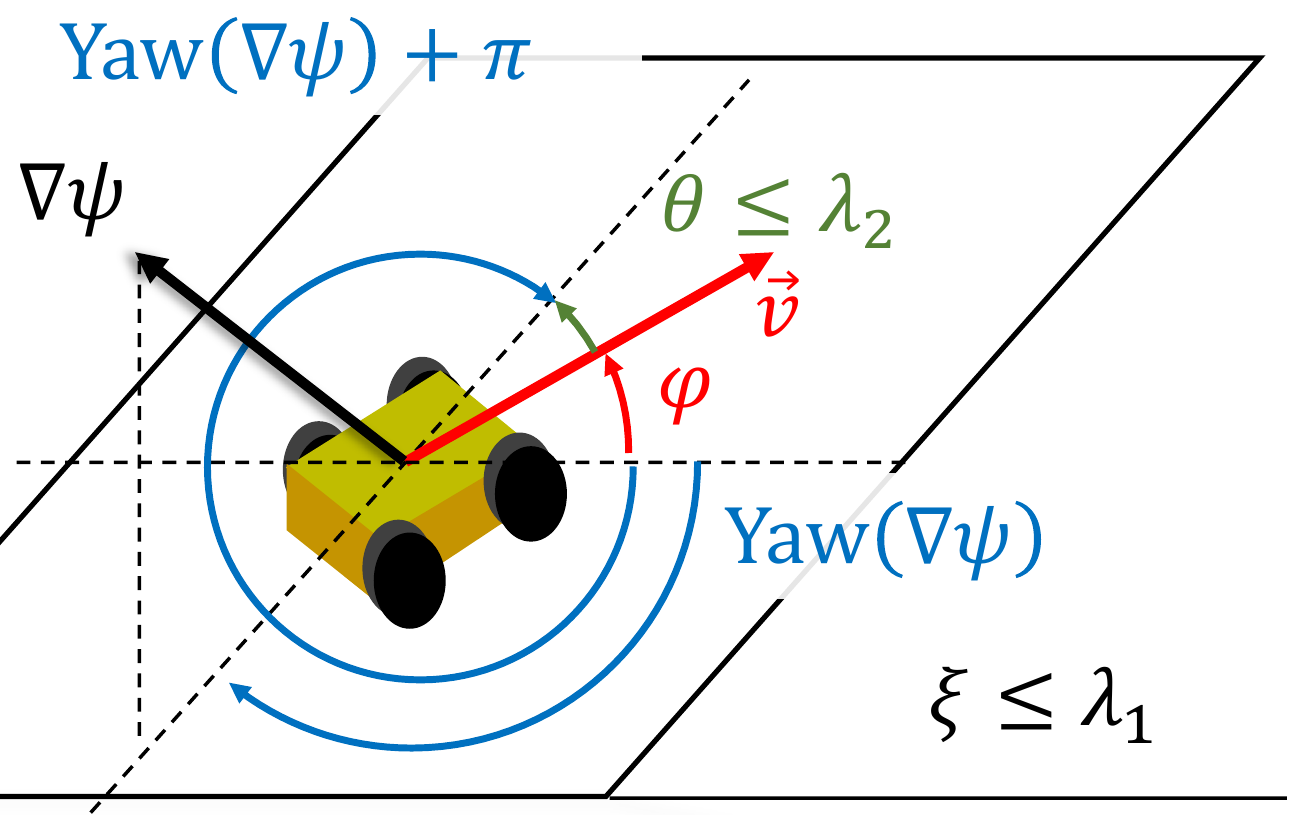}
    \caption{ \label{fig:risk_constraint_ab}}
\end{subfigure}
\hfill
\begin{subfigure}{0.45\columnwidth}
    \includegraphics[trim = 0mm 0mm 0mm 0mm, clip, width=1\columnwidth]{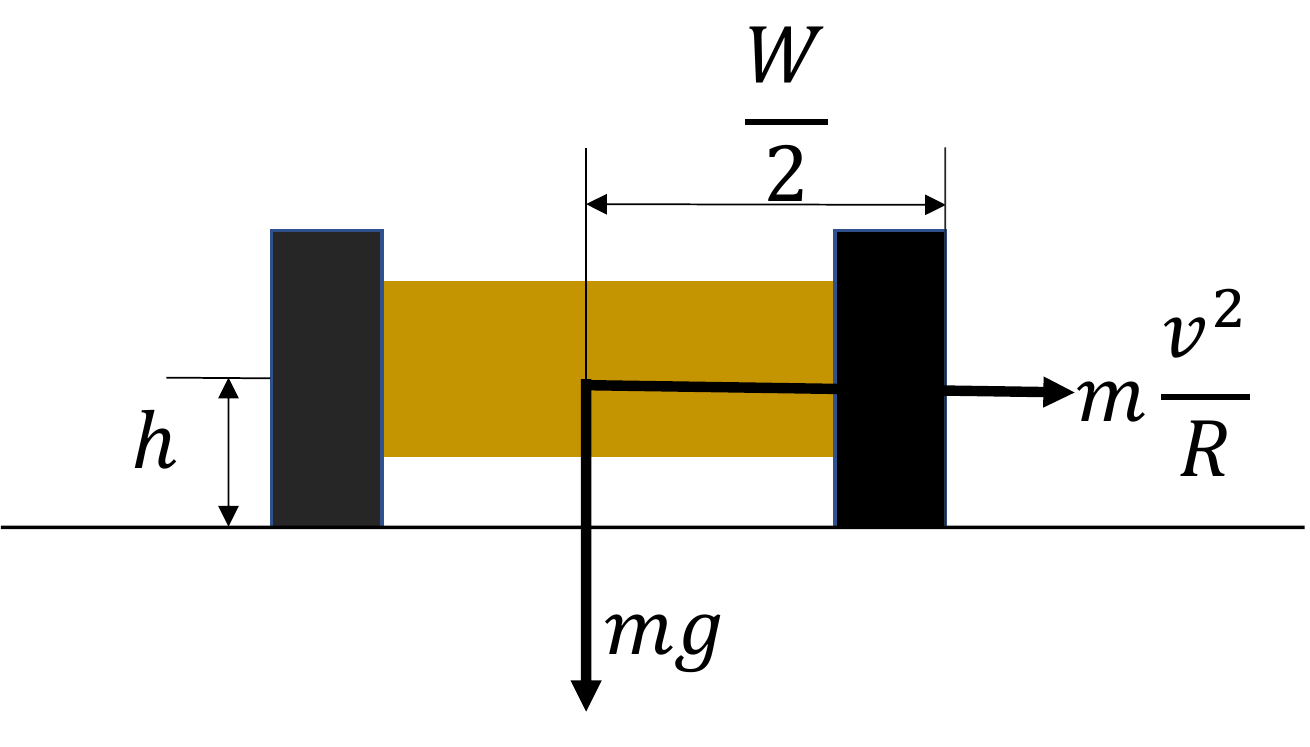}
    \caption{ \label{fig:risk_constraint_c}}
\end{subfigure}
\caption{ (a) describes risk constraints for steep slopes and turning on slopes. (b) describes the risk constraint for sharp turns at high speed \label{fig:risk_constraint}}
\end{figure}

\subsubsection{Steep slopes} Since the input costmap to the planner is 2D, we impose an explicit constraint to prevent the robot from climbing slopes that are too steep to be traversed safely. Mathematically, the robot maneuvers on the ground whose slope satisfies 
\begin{align}
\xi \leq \lambda_1,
\end{align}
where $\lambda_1$ is a threshold for safe slope, usually $45 \degree$.

\subsubsection{Turning on slopes} This is one of the most dangerous maneuvers the robot can undertake, as it has a high chance of high-centering the platform. Now, given a slope that is not too steep, the safest path for a 4 wheel ground robot is directly in the direction of the steepest ascent, to prevent turning on slopes at all times. We enforce this behavior by limiting the deviation of the robot's heading from the direction of the steepest ascent/descent at every point on the plan that involves climbing a slope. As shown in Fig. \ref{fig:risk_constraint_ab},

\begin{align}
    \theta \cdot \mathbb{1} (\xi \geq \xi_{\text{min}}) \leq \lambda_2,
\end{align}

where $\xi_{\text{min}}$ is usually $15 \degree$, $\lambda_2$ is usually $30 \degree$, and 

\begin{align} \theta = 
    \min\{\lvert \varphi - \text{Yaw}(\nabla \psi)  \rvert, \lvert \varphi - \pi - \text{Yaw}(\nabla \psi)  \rvert\}, 
\end{align}
where $\varphi$ is the robot's yaw angle, and $\text{Yaw}(\nabla \psi)$ is the yaw angle of the gradient vector of the slope. 

\subsubsection{Sharp turns at high speed} Due to the fast speeds and large sizes of many robots used in such applications, all turns must be subjected to tipping constraints. This is done by simply checking the required centripetal acceleration for the turning radius commanded and ensuring it falls within the limits of what the platform's geometry allows without tipping. Balancing torques in Fig. \ref{fig:risk_constraint_c} gives us
\begin{align}
    \frac{v^2}{R} \leq \lambda_3 \frac{Wg}{2h} .
\end{align}

where $v$ is the commanded linear velocity of the robot, $R$ is the instantaneous radius of curvature of the commanded turn, $W$ and $h$ are the platform's wheelbase and ground clearance respectively, $g$ is gravitational acceleration, and $\lambda_3 \geq 1$ is an optional safety factor.

\section{Results}
\label{sec:Results}

The mapping pipeline above is used as the foundation for running both the baseline min-time cost function and the constrained cost function described above in a high-fidelity Unity-based simulation environment as well as on two different robots in different outdoor scenarios. It is desirable to have an unknown space induced by a traversable terrain feature and obstacles in this unknown space that challenge the planner's safety. The ideal terrain feature that can produce this scenario is a hill, and therefore the tests done below involve climbing hill slopes.

\subsection{Simulation}
\label{sec:Simulation}

\begin{figure*}[th]
\centering
\begin{tabular}{cccc}
    \raisebox{0.0425\textheight}{\includegraphics[trim = 0mm 0mm 0mm 0mm, clip, height=0.090\textheight]{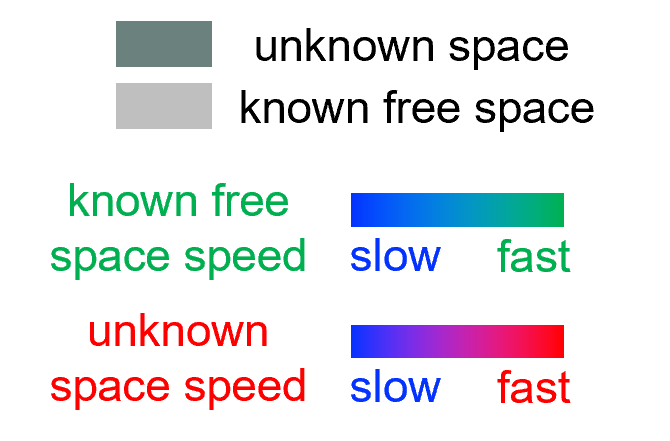}} &
    \includegraphics[trim = 0mm 0mm 0mm 0mm, clip, height=0.165\textheight]{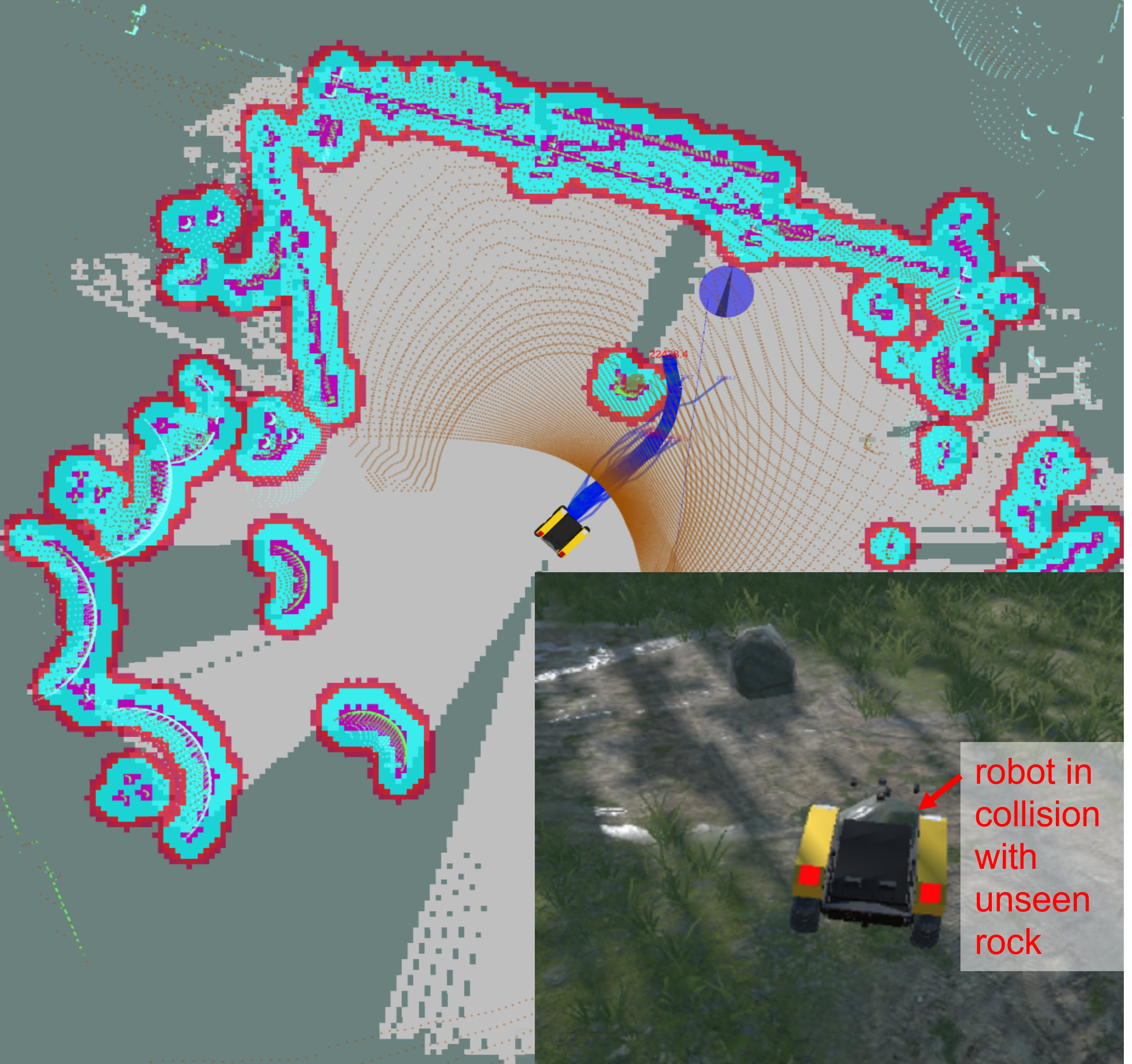} &
    \includegraphics[trim = 0mm 0mm 0mm 0mm, clip, height=0.165\textheight]{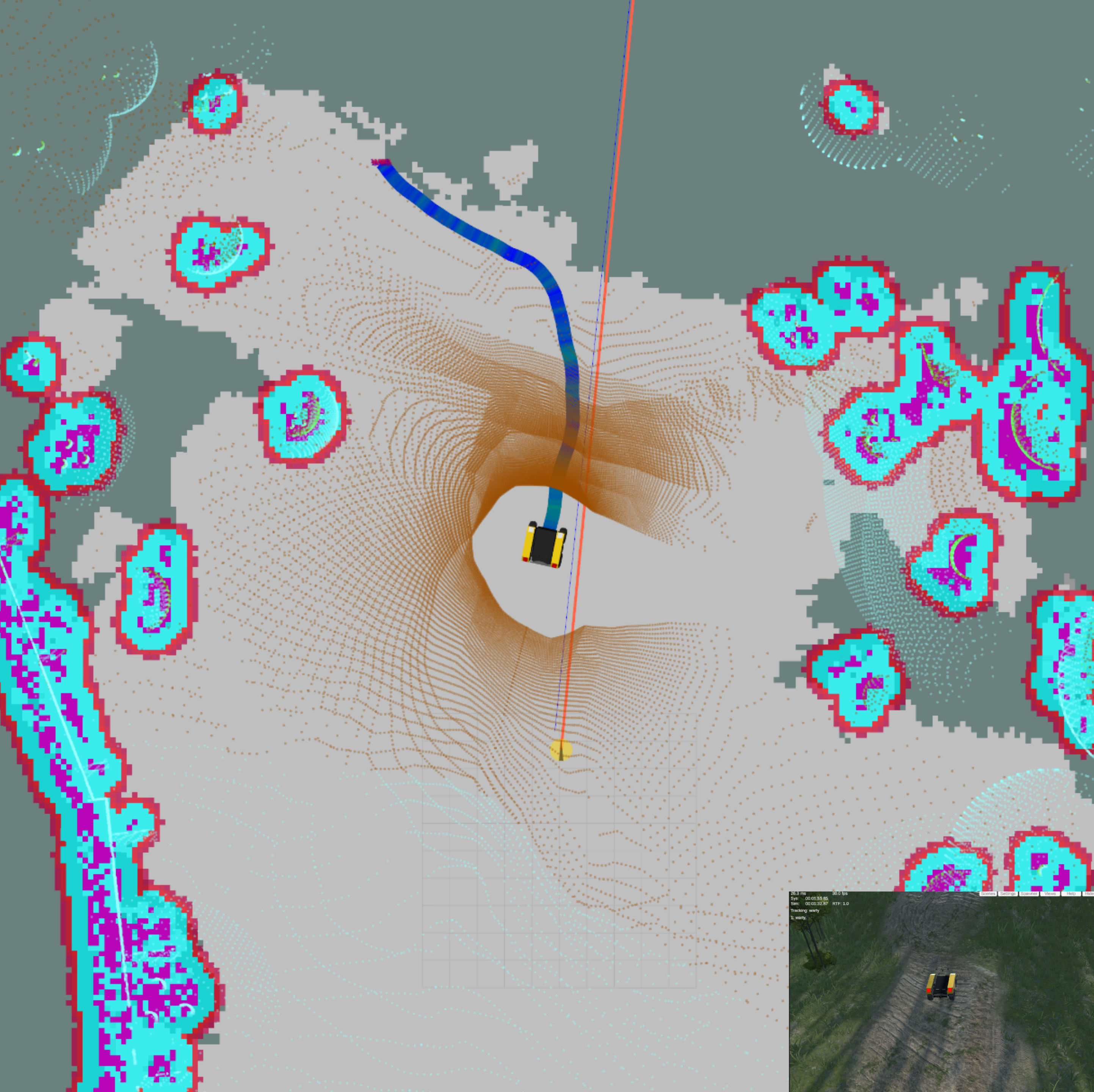} &
    \includegraphics[trim = 0mm 0mm 0mm 0mm, clip, height=0.165\textheight]{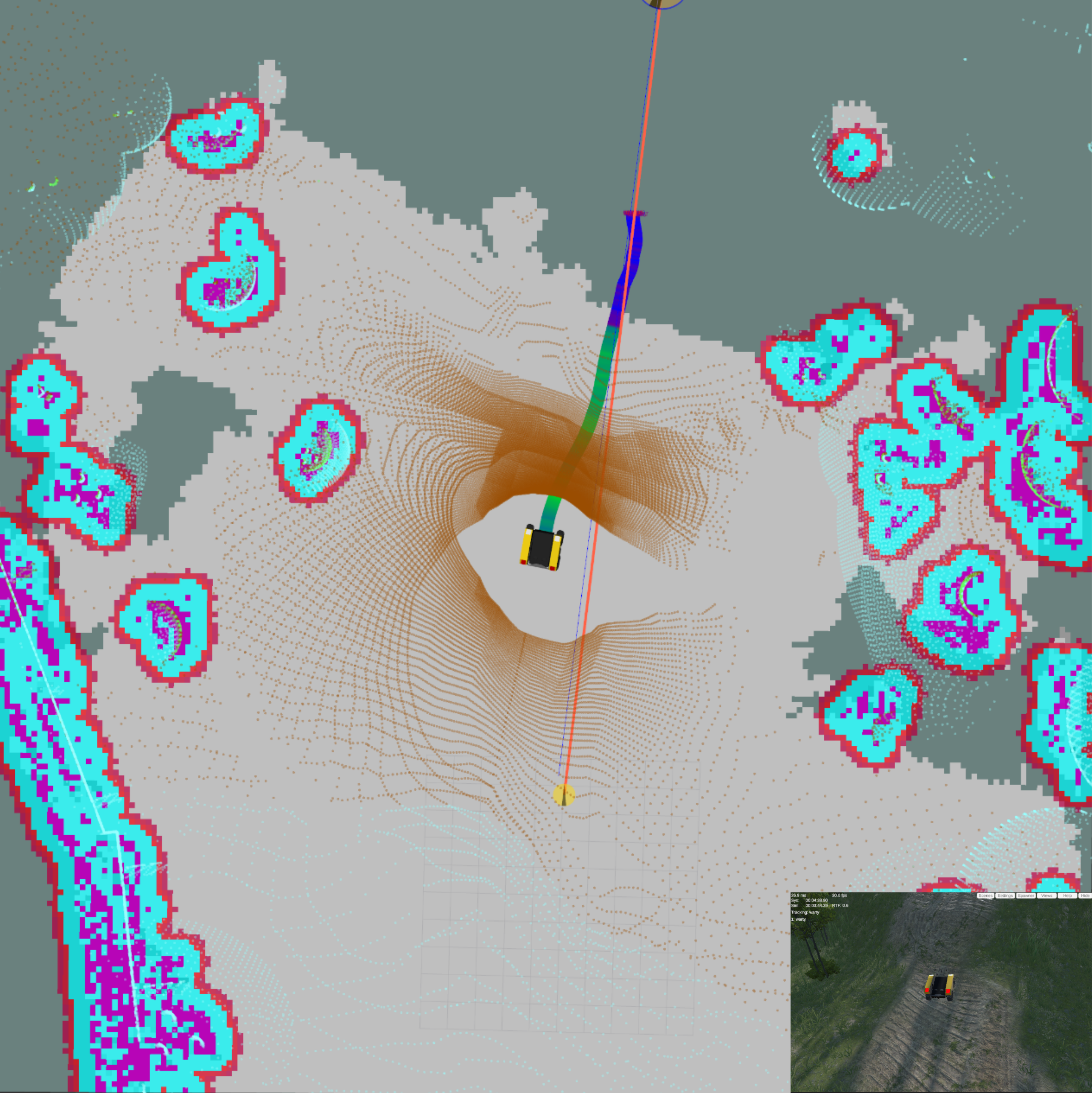} \\
    (a) & (b) & (c) & (d)
\end{tabular}
\caption{Simulation results: (a) The legend for known/unknown map regions and plan speeds used throughout this paper. (b) Pipeline 2: Robot collides with rocks beyond the hilltop 84\% of the time. (c) Pipeline 3: No collisions with rocks, but local plan curves on the hill, which leads to sliding, tipping, and slow speeds (more blue than green) in known free space. (d) RAMP with $\alpha = 60$: variability in horizon maximizes speed in known free space (solid green rollout in free space while mostly blue rollout in unknown space), and 3D risk constraints prevent turning on slopes.\label{fig:sim_results}}
\end{figure*}




\begin{figure}[th]
    \centering
\includegraphics[width=0.7\columnwidth]{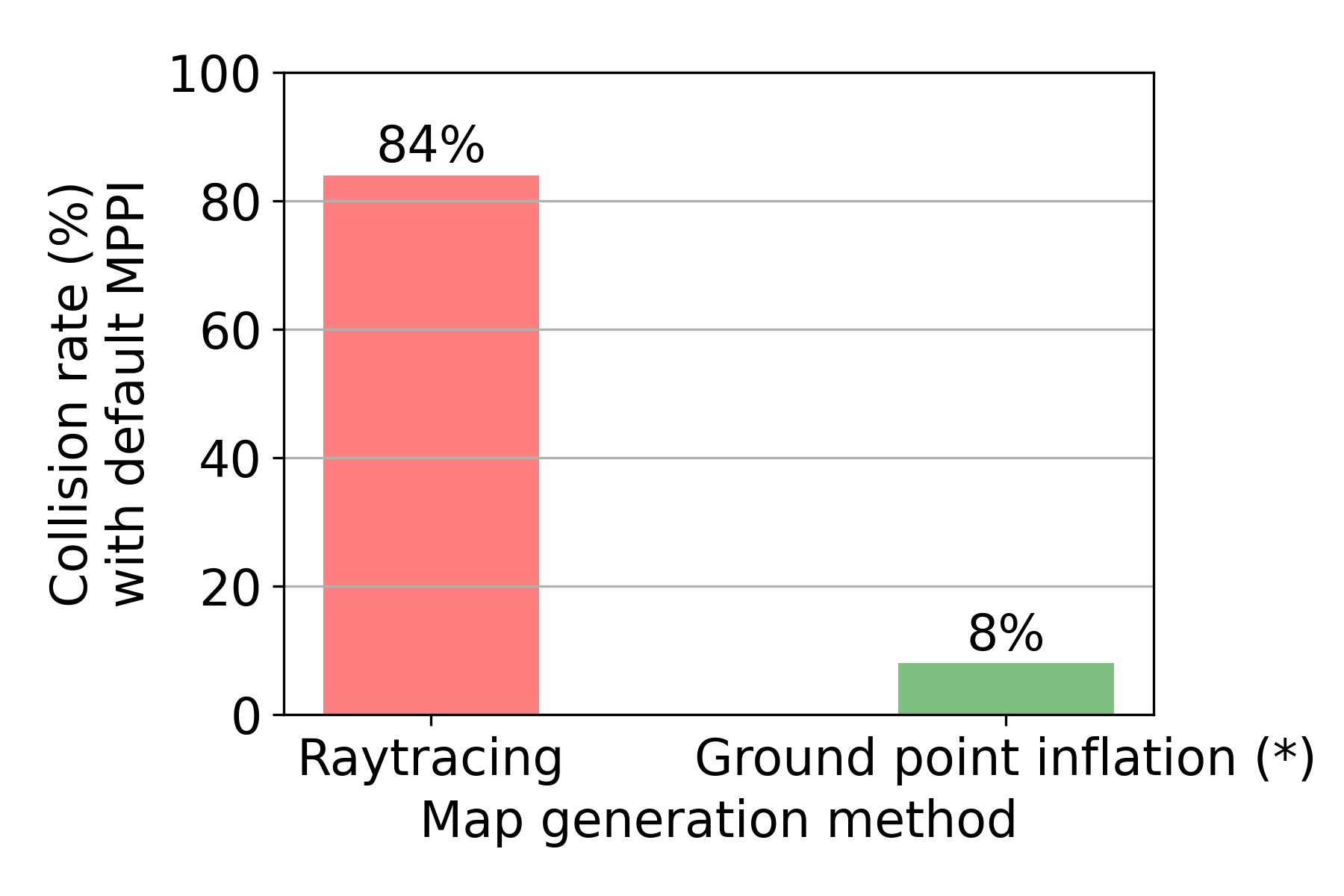}
    \caption{When running default MPPI with unknown space treated as an obstacle, there is a clear benefit to using ground point inflation over raytracing for map generation. Raytracing prematurely clears the hilltop as known free space and results in the robot frequently colliding with the rocks beyond.\label{fig:map_collision_rate_plot}}
\end{figure}

\begin{figure}[th]
\centering
 \includegraphics[width=0.8\columnwidth]{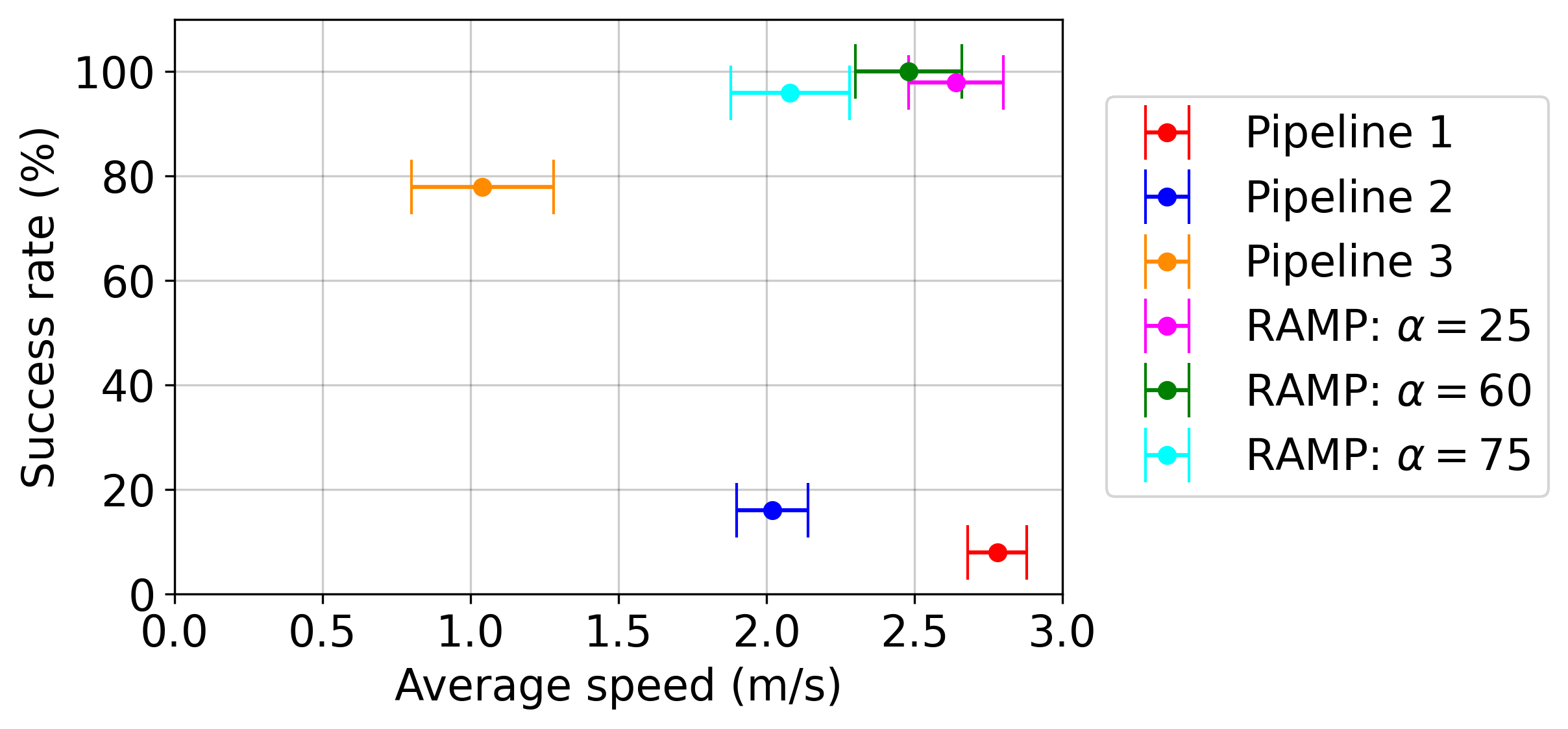}
\caption{Out of the 4 pipelines tested, only RAMP is able to achieve a full 100\% success rate at high speeds, thereby ensuring safety without sacrificing performance. \label{fig:success_vs_speed_plot}}
\end{figure}

The simulation environment chosen is indicated in Fig. \ref{fig:Map_with_without_memory} (a). It consists of flat terrain which quickly leads into a moderately steep hill with a slope of about \SI{30}{\degree}. At the top of this hill, there is flat land followed by tall metallic structures that are detectable by LiDAR from the base of the hill. For the purpose of the experiments, rocks that are big enough to block the robot are spawned at random locations on the flat land in each trial. The robot used in this environment is a Clearpath Warthog, which can achieve a top speed of \SI{5}{\m/\s}, which is also set as the maximum commanded speed for these experiments. All simulations were done in real-time using a desktop computer with a 3.6 GHz i7 CPU and an RTX2080 Super GPU.

\begin{table}[]
\centering
\caption{Overview of simulation tests}
\label{tab:pipelines}
\resizebox{\columnwidth}{!}{%
\begin{tblr}{
  hline{1,6} = {-}{0.08em},
  hline{2} = {-}{0.05em},
}
                     & \textbf{Mapping method} & \textbf{Planner}      & \textbf{Unknown Space} \\
\textbf{Pipeline 1}  & Raytracing              & Default MPPI          & Free                             \\
\textbf{Pipeline 2}  & Raytracing              & Default MPPI          & Impenetrable                             \\
\textbf{Pipeline 3}  & Ground point inflation  & Default MPPI          & Impenetrable                     \\
{\color[HTML]{FE0000} \textbf{RAMP}} & Ground point inflation  & Variable-horizon MPPI & 3D-aware risk                    
\end{tblr}
}
\end{table}

For each trial, the robot starts on flat land some distance away from the base of the hill and is given a fixed location (same for each trial) near the back of the hilltop as the goal. 50 such trials were conducted with each of the four pipelines described in Table \ref{tab:pipelines}. In all tests, the planning horizon $T$ was set to 10 seconds and the planner frequency was maintained at 20 Hz. For RAMP, three such sets of trials were done, with $\alpha$ set to 25, 60 and 75 to see how RAMP performs when unknown space caution is lowered or raised.

The average speed (for successful trials) and success and collision rates are measured, and qualitative plan behavior is observed throughout. Here, the collision rate is defined as the fraction of trials that hit rocks on the hilltop, and the success rate is defined as the fraction of trials that reach the goal. (inability to reach the goal can arise from either hitting a rock or tipping due to turning on a slope) This leads to the following results:
\begin{itemize}
    \item Pipeline 1 fails almost every time due to heading up the hill at high speeds and crashing into a rock at the top.
    \item Pipeline 2 also fails 84\% of the time, despite the high unknown space cost, due to the raytraced map not representing unknown space accurately and MPPI planning a high-speed trajectory up the slope. This is shown in Fig. \ref{fig:sim_results}b. It should be noted that Pipelines 1 and 2 have high average speeds in the few trials where they are successful due to the planner commanding a straight line trajectory at full speed up the hill. 
    \item Pipeline 3 does not hit rocks due to the accurate representation of unknown space by the inflated ground point map. This steep improvement in collision frequency is denoted in Fig. \ref{fig:map_collision_rate_plot}, showing the safety benefits of an inflated ground point map.
    \item However, Pipeline 3 has a success rate of only \SI{}{\%}, despite its low collision rate. This is because the fixed horizon of Pipeline 3 forces the long \SI{10}{\second} rollout to fit in a limited length of known free space. MPPI compensates for this by often introducing curvature in the local plan, as well as very low robot speed, as seen in Fig. \ref{fig:sim_results}c. This leads to a lot of wheel slip, sliding down the slope, and sometimes leading to the robot tipping over.
    \item The drawbacks of Pipeline 3, arising from both turning on slope and inability to adjust plan horizon based on the distance to unknown space, are fixed by RAMP. As shown in Fig. \ref{fig:sim_results}d, RAMP's plans with $\alpha = 60$ are not only straight up the hill but also fast in known space while slowing down in unknown space, achieving the desired 100\% success rate. Additionally, $\alpha$ can be increased to 75 to reduce the speed in unknown space further, which reduces average speed while maintaining a nearly perfect 96\% success rate, or $\alpha$ can be reduced to 25 to get a small increase in average speed and go faster in unknown space, at a slight reduction in the success rate to 98\%.
\end{itemize}

These results are summarized in Fig. \ref{fig:success_vs_speed_plot}, which shows that RAMP breaks the tradeoff barrier between success rate and high speed for challenging off-road environments.

\begin{figure*}[th]
\centering
\begin{tabular}{ccc}
    \includegraphics[trim = 0mm 0mm 0mm 0mm, clip, height=0.180\textheight]{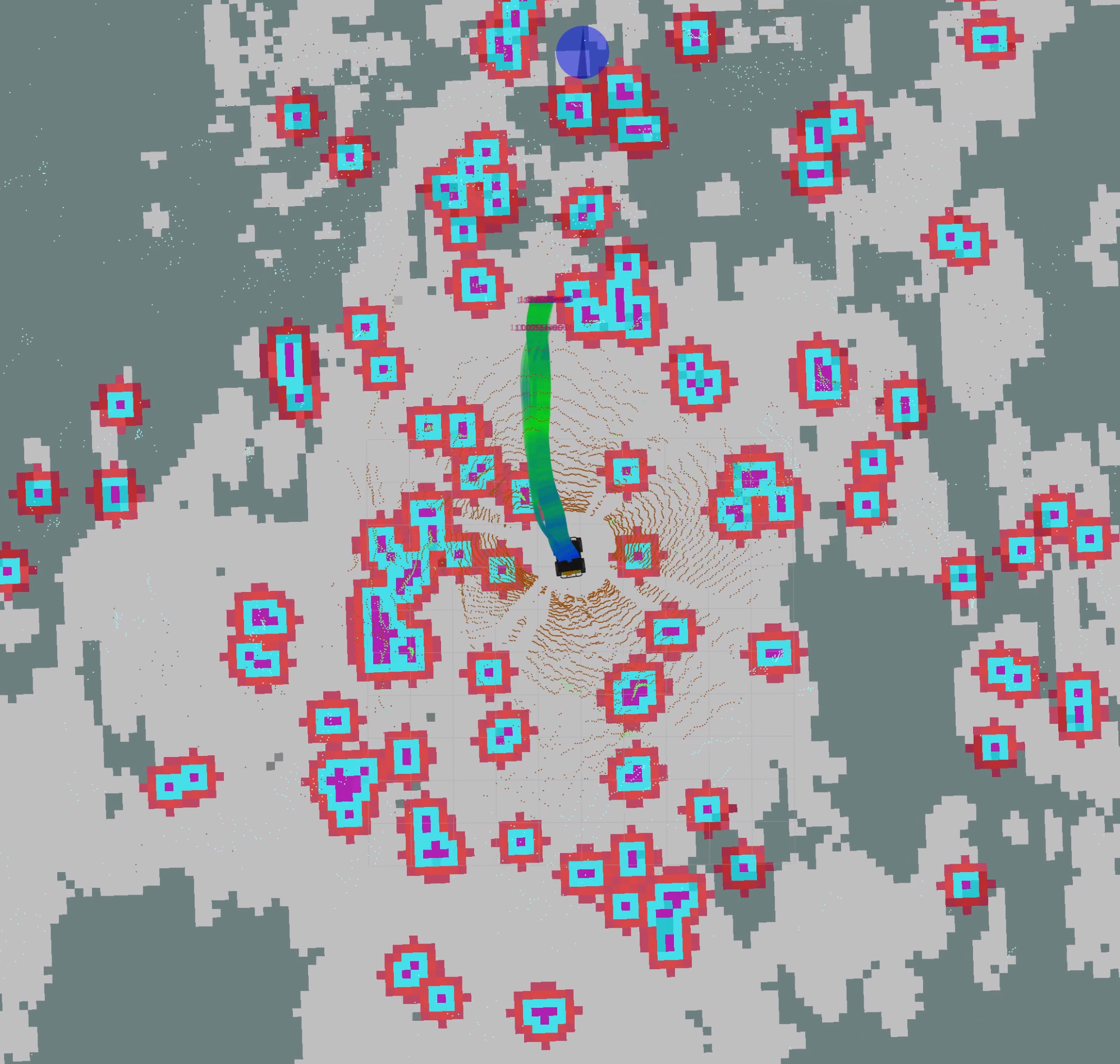} &
    \includegraphics[trim = 0mm 0mm 0mm 0mm, clip, height=0.180\textheight]{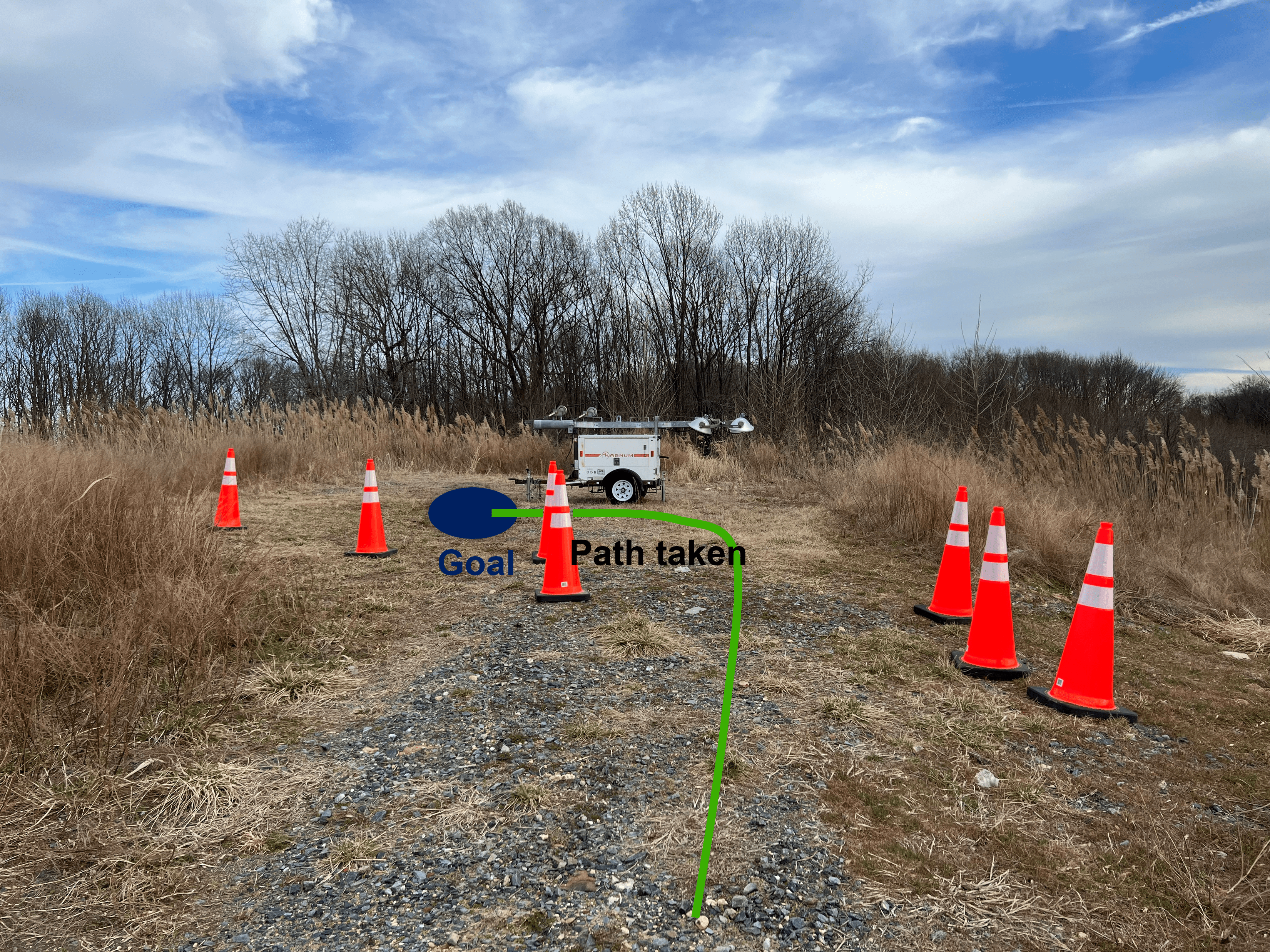} &
    \includegraphics[trim = 0mm 0mm 0mm 0mm, clip, height=0.180\textheight]{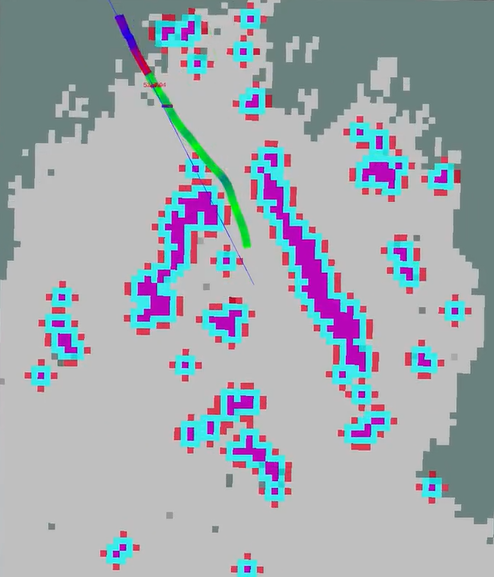} \\
    (a) & (b) & (c)
\end{tabular}
\caption{Hardware results: (a) The Husky was able to climb the hill fairly easily, RAMP's benefits are more visible on higher speed platforms. (b) The obstacles and goal/path for the Warthog experiment after clearing the hilltop. (c) The Warthog was able to climb up the hill fast yet safely, as shown by the slowdowns (blue line) in unknown space.\label{fig:hw_results}}
\end{figure*}

\subsection{Hardware Demonstrations}
\label{sec:HardwareDemonstrations}
RAMP was successfully deployed on two distinct platforms in two environments. First, we implemented RAMP on a Clearpath Husky robot, achieving a maximum speed of \SI{1}{\m/\s} to achieve real-time safe navigation at the platform's top speed. (planner frequency of 20 Hz for a \SI{10}{\second} planning horizon) This was done in the environment shown in Fig. \ref{fig:envs}a: a hill in a forest with about 30$\degree$ of slope. The main computer on the robot was equipped with a 2.8 GHz i7 CPU and RTX2060 GPU. The results can be summarized as follows:

\begin{itemize}
    \item The robot successfully climbed straight up the hill at full speed. The state of the local map and plan for this hill is shown in Fig. \ref{fig:hw_results}a.
    \item It was also made clear that the advantages of RAMP only become readily apparent at high speeds, since the long stopping distances involved in those scales necessitate the improvements that RAMP brings.
\end{itemize}


Next, we tested RAMP on a Clearpath Warthog robot, achieving a maximum speed of \SI{3}{\m/\s}. The platform's top speed is \SI{5}{\m/\s}, however we restricted it to \SI{3}{\m/\s} to mitgate the chance of a catastrophic failure. This was done in the environment shown in Fig. \ref{fig:envs}b: a 25$\degree$ hill with tall grass on either side, and numerous traffic cone obstacles beyond the hilltop for the robot to navigate around as shown in Fig. \ref{fig:hw_results}b. The main computer on the robot was equipped with two 2.8 GHz i7 CPU and two Tesla T4 GPUs. The results can be summarized as follows:

\begin{itemize}
    \item The robot successfully climbed straight up the hill at the max commanded speed of \SI{3}{\m/\s}. The state of the local map and plan for this hill is shown in Fig. \ref{fig:hw_results}c.
    \item There was a clear slowdown in the local plan at the edge of unknown space, due to the higher speed of the platform.
    \item No obstacles were hit and the robot was able to reach its goal safely, even though some obstacles were obscured due to the slope and needed to be avoided quickly to reach the goal.
\end{itemize}



\section{Conclusions and Future Work}
\label{sec:Conclusion}

This paper proposes a Risk-Aware Mapping and Planning (RAMP) pipeline for fast off-road ground robot navigation in unmapped 3D terrain. RAMP introduces a ground-plane support based map generation method by inflating classified ground points from LiDAR, combined with persistent memory for increasing map density. This map is then used by a novel MPPI-based planner, which introduces multiple layers of horizon variability to maximize speed in known free space while allowing for tunable speed in unknown space. When compared to default MPPI with different levels of mapping and unknown space costing improvements, only RAMP delivers high success rates at high speed, and is interfaced with 3D aware risk constraints to allow for safe and fast navigation up hills in both simulation and on hardware.

In the near future, we plan to further develop the risk heuristic for unknown space beyond simply average speed, and distinguish between different kinds of unknown space. We also plan to address the risk of map smear introduced by the persistence of the proposed spatial memory. Additionally, we plan to scale up by testing RAMP at higher speeds of \SI{5}{\m/\s} and beyond, and in more challenging terrain such as a steep ravine with nearly $45 \degree$ slopes. We also plan to use a better LiDAR classifier such as Cylinder3D \cite{zhou2020cylinder3d} to get ground points for map generation, since the hardware tests showed that geometric classifiers struggle with tall grass.
%

%





\section*{ACKNOWLEDGMENT}
Research was sponsored by the Army Research Office under Cooperative Agreement Number
W911NF-21-2-0150 as well as under the DARPA RACER program. The views and conclusions contained
in this document are those of the authors and should not
be interpreted as representing the official policies, either
expressed or implied, of the Army Research Office or the
U.S. Government. The U.S. Government is authorized to
reproduce and distribute reprints for Government purposes
notwithstanding any copyright notation herein.


\bibliographystyle{IEEEtran}
\bibliography{references}

\begin{thebibliography}{10}
\providecommand{\url}[1]{#1}
\csname url@samestyle\endcsname
\providecommand{\newblock}{\relax}
\providecommand{\bibinfo}[2]{#2}
\providecommand{\BIBentrySTDinterwordspacing}{\spaceskip=0pt\relax}
\providecommand{\BIBentryALTinterwordstretchfactor}{4}
\providecommand{\BIBentryALTinterwordspacing}{\spaceskip=\fontdimen2\font plus
\BIBentryALTinterwordstretchfactor\fontdimen3\font minus
  \fontdimen4\font\relax}
\providecommand{\BIBforeignlanguage}[2]{{%
\expandafter\ifx\csname l@#1\endcsname\relax
\typeout{** WARNING: IEEEtran.bst: No hyphenation pattern has been}%
\typeout{** loaded for the language `#1'. Using the pattern for}%
\typeout{** the default language instead.}%
\else
\language=\csname l@#1\endcsname
\fi
#2}}
\providecommand{\BIBdecl}{\relax}
\BIBdecl

\bibitem{bares1989ambler}
J.~Bares, M.~Hebert, T.~Kanade, E.~Krotkov, T.~Mitchell, R.~Simmons, and
  W.~Whittaker, ``Ambler: An autonomous rover for planetary exploration,''
  \emph{Computer}, vol.~22, no.~6, pp. 18--26, 1989.

\bibitem{massari2004autonomous}
M.~Massari, G.~Giardini, and F.~Bernelli-Zazzera, ``Autonomous navigation
  system for planetary exploration rover based on artificial potential
  fields,'' in \emph{Proceedings of Dynamics and Control of Systems and
  Structures in Space (DCSSS) 6th Conference}, 2004, pp. 153--162.

\bibitem{voosen2018nasa}
P.~Voosen, ``Nasa curiosity rover hits organic pay dirt on mars,'' 2018.

\bibitem{kantor2003distributed}
G.~Kantor, S.~Singh, R.~Peterson, D.~Rus, A.~Das, V.~Kumar, G.~Pereira, and
  J.~Spletzer, ``Distributed search and rescue with robot and sensor teams,''
  in \emph{Field and Service Robotics}.\hskip 1em plus 0.5em minus 0.4em\relax
  Springer, 2003, pp. 529--538.

\bibitem{yamauchi2004packbot}
B.~M. Yamauchi, ``Packbot: a versatile platform for military robotics,'' in
  \emph{Unmanned ground vehicle technology VI}, vol. 5422.\hskip 1em plus 0.5em
  minus 0.4em\relax SPIE, 2004, pp. 228--237.

\bibitem{elfes1989using}
A.~Elfes, ``Using occupancy grids for mobile robot perception and navigation,''
  \emph{Computer}, vol.~22, no.~6, pp. 46--57, 1989.

\bibitem{fankhauser2016gridmap}
\BIBentryALTinterwordspacing
P.~Fankhauser and M.~Hutter, ``{A Universal Grid Map Library: Implementation
  and Use Case for Rough Terrain Navigation},'' in \emph{Robot Operating System
  (ROS) – The Complete Reference (Volume 1)}, A.~Koubaa, Ed.\hskip 1em plus
  0.5em minus 0.4em\relax Springer, 2016, ch.~5. [Online]. Available:
  \url{http://www.springer.com/de/book/9783319260525}
\BIBentrySTDinterwordspacing

\bibitem{hornung2013octomap}
A.~Hornung, K.~M. Wurm, M.~Bennewitz, C.~Stachniss, and W.~Burgard, ``Octomap:
  An efficient probabilistic 3d mapping framework based on octrees,''
  \emph{Autonomous robots}, vol.~34, no.~3, pp. 189--206, 2013.

\bibitem{stuckler2014multi}
J.~St{\"u}ckler and S.~Behnke, ``Multi-resolution surfel maps for efficient
  dense 3d modeling and tracking,'' \emph{Journal of Visual Communication and
  Image Representation}, vol.~25, no.~1, pp. 137--147, 2014.

\bibitem{wiemann2019file}
T.~Wiemann, F.~Igelbrink, S.~P{\"u}tz, and J.~Hertzberg, ``A file structure and
  reference data set for high resolution hyperspectral 3d point clouds,''
  \emph{IFAC-PapersOnLine}, vol.~52, no.~8, pp. 403--408, 2019.

\bibitem{oleynikova2016signed}
H.~Oleynikova, A.~Millane, Z.~Taylor, E.~Galceran, J.~Nieto, and R.~Siegwart,
  ``Signed distance fields: A natural representation for both mapping and
  planning,'' in \emph{RSS 2016 workshop: geometry and beyond-representations,
  physics, and scene understanding for robotics}.\hskip 1em plus 0.5em minus
  0.4em\relax University of Michigan, 2016.

\bibitem{tordesillas2021faster}
J.~Tordesillas, B.~T. Lopez, M.~Everett, and J.~P. How, ``Faster: Fast and safe
  trajectory planner for navigation in unknown environments,'' \emph{IEEE
  Transactions on Robotics}, vol.~38, no.~2, pp. 922--938, 2021.

\bibitem{fridovich2019safely}
D.~Fridovich-Keil, J.~F. Fisac, and C.~J. Tomlin, ``Safely probabilistically
  complete real-time planning and exploration in unknown environments,'' in
  \emph{2019 International Conference on Robotics and Automation (ICRA)}.\hskip
  1em plus 0.5em minus 0.4em\relax IEEE, 2019, pp. 7470--7476.

\bibitem{kousik2020safe}
S.~Kousik, B.~Zhang, P.~Zhao, and R.~Vasudevan, ``Safe, optimal, real-time
  trajectory planning with a parallel constrained bernstein algorithm,''
  \emph{IEEE Transactions on Robotics}, vol.~37, no.~3, pp. 815--830, 2020.

\bibitem{janson2018safe}
L.~Janson, T.~Hu, and M.~Pavone, ``Safe motion planning in unknown
  environments: Optimality benchmarks and tractable policies,'' \emph{arXiv
  preprint arXiv:1804.05804}, 2018.

\bibitem{cai2022risk}
X.~Cai, M.~Everett, J.~Fink, and J.~P. How, ``Risk-aware off-road navigation
  via a learned speed distribution map,'' \emph{arXiv preprint
  arXiv:2203.13429}, 2022.

\bibitem{shekhar2012robust}
R.~C. Shekhar and J.~M. Maciejowski, ``Robust variable horizon mpc with move
  blocking,'' \emph{Systems \& Control Letters}, vol.~61, no.~4, pp. 587--594,
  2012.

\bibitem{williams2017model}
G.~Williams, A.~Aldrich, and E.~A. Theodorou, ``Model predictive path integral
  control: From theory to parallel computation,'' \emph{Journal of Guidance,
  Control, and Dynamics}, vol.~40, no.~2, pp. 344--357, 2017.

\bibitem{zhou2020cylinder3d}
H.~Zhou, X.~Zhu, X.~Song, Y.~Ma, Z.~Wang, H.~Li, and D.~Lin, ``Cylinder3d: An
  effective 3d framework for driving-scene lidar semantic segmentation,''
  \emph{arXiv preprint arXiv:2008.01550}, 2020.

\bibitem{aksoy2020salsanet}
E.~E. Aksoy, S.~Baci, and S.~Cavdar, ``Salsanet: Fast road and vehicle
  segmentation in lidar point clouds for autonomous driving,'' in \emph{2020
  IEEE intelligent vehicles symposium (IV)}.\hskip 1em plus 0.5em minus
  0.4em\relax IEEE, 2020, pp. 926--932.

\end{thebibliography}

\end{document}